\documentclass[conference]{IEEEtran}

\usepackage{enumerate}
\usepackage{graphicx}
\usepackage{amsmath,epsfig,amssymb,mathrsfs}
\usepackage{algorithmic}
\usepackage{cite}
\usepackage{amsthm}
\usepackage{arydshln}
\usepackage{amsopn}
\usepackage{color}
\usepackage{verbatim}
\usepackage{amsopn}
\usepackage{amssymb}
\usepackage{algorithm}
\usepackage{enumerate}
\usepackage{subcaption}
\usepackage{float}
\floatstyle{plaintop}
\restylefloat{table}

\newcommand{\bx}{\mathbf{x}}
\newcommand{\by}{\mathbf{y}}
\newcommand{\bR}{\mathbb{R}}

\newcommand{\bPsi}{\boldsymbol \Psi}
\newcommand{\bPhi}{\boldsymbol \Phi}
\IEEEoverridecommandlockouts

\begin{document}
\title{A Deep Learning Approach \\ to Structured Signal Recovery}
\author{
\IEEEauthorblockN{Ali Mousavi, Ankit B. Patel, Richard G. Baraniuk  }
\IEEEauthorblockA{
\\Department of Electrical and Computer Engineering \\
Rice University\\
Houston, TX 77005\\
}
\thanks{ This work was supported by NSF CCF-0926127, CCF-1117939; DARPA/ONR N66001-11-C-4092 and N66001-11-1-4090; ONR N00014-10-1-0989, and N00014-11-1-0714; ARO MURI W911NF-09-1-0383.

Email: \{ali.mousavi, abp4, richb\} @rice.edu }
}
\maketitle
\begin{abstract}
In this paper, we develop a new framework for sensing and recovering structured signals. In contrast to compressive sensing (CS) systems that employ linear measurements, sparse representations, and computationally complex convex/greedy algorithms, we introduce a deep learning framework that supports both linear and mildly nonlinear measurements, that learns a structured representation from training data, and that efficiently computes a signal estimate. In particular, we apply a stacked denoising autoencoder (SDA), as an unsupervised feature learner. SDA enables us to capture statistical dependencies between the different elements of certain signals and improve signal recovery performance as compared to the CS approach.\end{abstract}

\section{Introduction}
\subsection{Motivation}
An inverse problem that occurs in a number of important applications involves recovering a signal $\bx \in \bR^N$ from a set of under-sampled measurements. This problem is formulated as recovering $\bx \in \bR^N$ from $\by = \boldsymbol \Gamma(\bx)$ where $\boldsymbol \Gamma(.):\bR^N \to \bR^M$ could be either a linear or non-linear function while $M\ll N$. Since this problem is ill-posed in general, one is able to recover $\bx$ given $\by$ and $\boldsymbol \Gamma(.)$ only if $\bx$ has some type of structure such that by applying $\boldsymbol \Gamma(.)$ its dimensionality can be reduced from $N$ without losing information. Many configurations for $\bx$ and $\boldsymbol \Gamma(.)$ have been explored in the literature for this problem; however, one of the most useful ones is to have a sparse signal $\bx$ and a linear $\boldsymbol \Gamma(.)$, i.e., $\by = \boldsymbol \Gamma(\bx)=\bPhi \bx$  . Compressive sensing (CS) \cite{donoho2006compressed,baraniuk2007compressive, candes2006compressive} is a field that tries to solve this linear inverse problem in case that $\bx$ has a sparse representation, i.e., there exists   an $N\times N$ basis matrix $\boldsymbol \Psi = [\boldsymbol \psi_1|\boldsymbol \psi_2|\ldots|\boldsymbol \psi_N]$ such that $\mathbf{x} = \boldsymbol \Psi \mathbf{s}$ and only $K \ll N$ of the coefficients $\mathbf{s}$ are nonzero. Therefore, CS is mainly concerned with the problem of recovering a $K$-sparse signal $\mathbf{x} \in \mathbb{R}^N$ from a set of under-sampled linear measurements, i.e., from  $M \ll N$ measurements acquired via $\mathbf{y} = \boldsymbol \Phi \mathbf{x} =  \boldsymbol \Phi  \boldsymbol \Psi \mathbf{s}$, where $\mathbf{y} \in \mathbb{R}^M$ is the measurement vector and $\boldsymbol \Phi \in \mathbb{R}^{M\times N}$ is the measurement matrix. 

The measurement vector formulation $\by = \boldsymbol \Gamma(\bx)$ suggests that one should answer the following questions to compressively acquire a signal: 
\begin{enumerate}[(i)]
\item How to recover the signal $\mathbf{x}$ from a given measurement vector $\mathbf{y}$ and operator $\boldsymbol \Gamma(.)$? 
\item How to design the measurement operator $\boldsymbol \Gamma(.)$? 
\item If we are concerned with any type of structure, How could we find a representation in which the signal $\mathbf{x}$ has that structure?
\end{enumerate}

In case of sparse $\bx$ and linear $\boldsymbol \Gamma(.)$, CS framework answers these three questions in the following way:
\begin{enumerate}[(i)]
\item Using methods from convex optimization or greedy algorithms.
\item Using linear random matrices as measurement matrices.
\item Using pre-specified set of transformations or data-dependent basis such as wavelets, frames, and dictionaries.
\end{enumerate}

Although there has been a considerable progress in CS and particularly in the answers of aforementioned questions, our goal is to go beyond the state-of-the-art results. We approach this goal by incorporating a deep learning framework into structured signal recovery. 

Deep learning is an emerging field mainly concerned with learning multiple levels of representation of data and coming up with higher levels of abstraction in it. In this paper we study the ability of deep neural networks to recover structured signals (in particular images) from their under-sampled random linear measurements. In other words, we study the performance of deep learning framework in recovering structured signals from their under-sampled measurements. 

The motivation for this work is the great success of deep architectures in image representation. In particular, Hinton et al. showed in \cite{hinton2006reducing} that one can achieve dimensionality reduction in high-dimensional data by training a multilayer neural network called autoencoder. We compare the performance of deep learning approach with state-of-the-art algorithms for solving the CS problem and show that deep architectures can help us to outperform their results at least in certain cases.

In the following three paragraphs, we briefly describe how deep learning provides new opportunities to attack questions (i), (ii), and (iii) mentioned above. We specifically compare these opportunities with the answers to these questions given by CS framework.

The first question is about recovering the original signal from measurement vector and matrix. In order to recover a sparse signal $\mathbf{x} \in \mathbb{R}^N$ from its corresponding measurement vector $\mathbf{y} \in \mathbb{R}^M$, one needs to seek for the sparsest signal $\mathbf{\hat{x}}$ that agrees with the measurement vector $\mathbf{y}$
\begin{align}\label{eq:l0}
\mathbf{\hat{x}} = \arg \min_{\mathbf{x'}} \|\mathbf{x'}\|_0 ~~ \text{s.t.} ~~ \mathbf{y}=\boldsymbol \Phi \mathbf{x'},
\end{align}  
where $\|.\|_0$ denotes the $\ell_0$-norm of a vector and counts the number of its nonzero elements. While it has been shown that by using only $\mathcal{O}(K)$ measurements this optimization can recover a $K$-sparse signal \cite{baron2005distributed}, solving \eqref{eq:l0} is an NP-hard problem. Therefore, researchers have replaced $\ell_0$-norm in \eqref{eq:l0} with its convex relaxation $\ell_1$-norm to convert \eqref{eq:l0} to a tractable and stable linear programming problem. 
This linear program can be solved either based on convex optimization methods \cite{candes2006near} or iterative greedy algorithms \cite{donoho2009message, needell2009cosamp, blumensath2009iterative, beck2009fast} that are generally first order methods and as a result are more suitable for high-dimensional problems. In this paper, we replace these algorithms, i.e., the convex optimization based approaches and greedy iterative algorithms converging usually in hundreds of iterations with a feed-forward deep neural network. We show that as a result of using a feed-forward deep neural network we do not need to solve the linear program to recover $\bx$ and hence we can have much faster signal recovery.

The second question is about designing measurement matrix. Traditional approaches in designing measurement matrix $\boldsymbol \Phi$ are based on focusing on desirable properties needed by $\boldsymbol \Phi$ to preserve information while doing dimensionality reduction, i.e., mapping $\mathbf{x} \in \mathbb{R}^N$ to $\mathbf{y} \in \mathbb{R}^M$ where $M \ll N$. One important property of measurement matrix that guarantees successful sparse signal recovery with very high probability is \textit{restricted isometry property} (RIP) \cite{candes2008restricted}. While checking whether or not a matrix has the RIP is an NP-Complete problem, random matrices whose elements are independent and identically distributed (i.i.d.) Gaussian or Bernoulli random variables, satisfy the RIP with very high probability given $M=\mathcal{O}(K\log(N/K))$. The main drawback of random measurements is that they are not optimally designed according to the signal under acquisition. Adaptive methods \cite{malloy2014near, haupt2009adaptive, haupt2012sequentially, haupt2009compressive} in which each measurement is designed based on the information obtained from previous measurements reduce uncertainty. The major problem with adaptive sequential measurements in CS is time complexity since each new measurement will depend on the information obtained from prior measurements. In this paper we show how deep neural networks can help us to adapt the measurements to the signal being under acquisition instead of taking random measurements and hence enhance the performance of the overall system.

Finally, the third question asks about finding a representation in which the original signal $\bx$ has a specific structure. In CS framework, one is concerned with finding a basis $\bPsi$ in which $\bx$ has a sparse representation. It is well known that $\bPsi$ could be chosen from a prespecified set of transformations. For example, natural images have sparse representation in wavelet basis or in the DCT domain \cite{mallat1999wavelet}. The main drawback of these prespecified bases is that they are handcrafted and as a result restrictive in capturing complex dependencies between different elements of a signal. More concretely, the main drawback of representing an image in wavelet domain is the assumption of independence between wavelet coefficients. This point has motivated researchers to develop models for capturing statistical dependencies in real-world signals \cite{crouse1998wavelet, de1997non}. However, these models are also handcrafted and hence do not necessarily capture more complex dependencies within a signal. The limitation in representation power of these prespecified transformations has lead researchers to seek for data-dependent basis, i.e., learning a transformation from a set of training examples \cite{aharon2006img}. Deep learning is a framework based on automating feature discovery and feature learning for many machine learning tasks. Accordingly, in this paper we use deep learning framework to automate the process of finding a representation for a class of signals being under acquisition. We show how the learned representation outperforms prespecified set of transformations. In particular, we focus on image data and show how deep neural networks outperform wavelet domain by providing a better representation to do dimensionality reduction.

We believe, to the best of our knowledge, that this paper is the first one trying to study structured signal recovery from  a set of under-sampled measurements by using deep learning framework. However, there have been several studies of using deep learning technique in solving the inverse problems. These studies have been focused on image denoising \cite{burger2012image}, removing noisy patterns from images \cite{eigen2013restoring}, and image super-resolution \cite{dong2014learning}. 

The rest of this paper is organized as follows: Section \ref{sec:arc} introduces the network architecture we have used to solve the structured signal recovery problem. In Section \ref{sec:theory} we discuss the probabilistic interpretation for using stacked denoising autoencoders (SDA) in solving the structured signal recovery problem. Section \ref{sec:simul} contains the simulation results. Finally, Section \ref{sec:conclu} includes the conclusion of the paper. 

\section{Stacked Denoising Autoencoders for Structured Signal Recovery}\label{sec:arc}
As we mentioned, natural images have sparse representation in wavelet basis or in the DCT domain. Therefore, they could be compressively acquired and reconstructed  using CS framework.  In this section we introduce our deep architecture for solving this CS recovery problem. Later in Section \ref{sec:simul} we compare the performance of the proposed method in this section with other state-of-the-art approaches from CS framework. 

We divide our solution into two different scenarios. First, we consider fixed linear measurements that is the traditional CS measurement paradigm. Second, we introduce a new measurement paradigm, namely nonlinear adaptive compressive measurements, inspired by neural networks architecture and capability. Later in Section \ref{sec:simul} we will show that incorporating nonlinearity in the measurements enhance the overall recovery performance.
\subsection{SDA + Linear Measurement Paradigm}\label{sec:SDAL}
In linear measurement paradigm, the measurement vector $\mathbf{y}$ is represented as $\mathbf{y}=\boldsymbol \Phi \mathbf{x}$, i.e., each $\mathbf{y}_i$ ($1 \leq i \leq M$) is a linear combination of $\mathbf{x}_j$s ($1\leq j \leq N$). We consider the typical supervised learning framework where our training set $\mathcal{D}_{\rm train}$ has $l$ pairs consisting of original signals and their corresponding measurements, i.e.,  $\mathcal{D}_{\rm train} = \{(\mathbf{y}^{(1)},\mathbf{x}^{(1)}),(\mathbf{y}^{(2)},\mathbf{x}^{(2)}),\ldots,(\mathbf{y}^{(l)},\mathbf{x}^{(l)})\}$. Based on this training set, we would like to learn a nonlinear mapping from a measurement vector $\mathbf{y}$ to its original signal $\mathbf{x}$. We then test the performance of the trained deep architecture on our test set $\mathcal{D}_{\rm test}$ where it has $s$ pairs consisting of original signals and their corresponding measurements, i.e.,  $\mathcal{D}_{\rm test} = \{(\mathbf{y}^{(1)},\mathbf{x}^{(1)}),(\mathbf{y}^{(2)},\mathbf{x}^{(2)}),\ldots,(\mathbf{y}^{(s)},\mathbf{x}^{(s)})\}$.
 
\begin{figure}[t!]
\begin{center}
\includegraphics[width= 7cm]{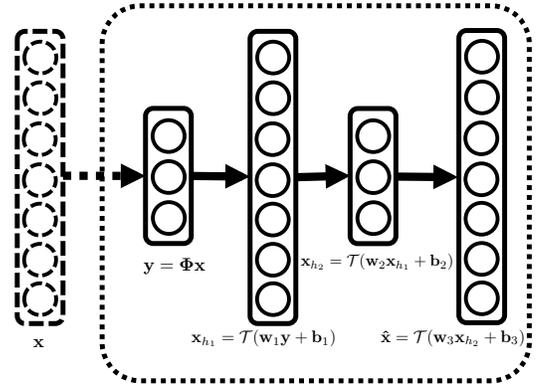}
\caption{Stacked denoising autoencoders (SDA) for recovering a sparse signal from its linear measurements. This is equivalent to having a 3-layer neural network fed with linear measurements of the original signal and try to reconstruct it.}
\label{fig:SDA_L}
\end{center}
\end{figure}

Among the traditional sparse recovery algorithms, the ones that are greedy and iterative perform faster than the ones that are based on convex optimization techniques such as linear programming. Each iteration in these greedy or iterative algorithms includes a matrix-vector multiplication which has the computational cost of $\mathcal{O}(MN)$. This fact was an inspiration for us to design the SDA architecture such that its recovery speed would be competitive with the existing fast iterative algorithms along with being similar to an iterative message passing algorithm. Therefore, each layer of the SDA used for sparse recovery either has input size of $N$ (the ambient dimension of the original signal) and output size of $M$ (the dimension of the measurement vector) or vice versa. We use a 3-layer SDA where each layer applies a nonlinearity to the affine transformation of its input. More formally, the first hidden layer receiving measurement vector as its input is formulated as
\begin{align}\label{eq:1stLayer}
\mathbf{x}_{h_1}=\mathcal{T}(\mathbf{W}_1\mathbf{y}+\mathbf{b}_1),
\end{align}
where $\mathbf{W}_1 \in \mathbf{R}^{N\times M}$ and $\mathbf{b}_1 \in \mathbf{R}^N$ are the weight matrix and bias vector of the first layer, respectively. $\mathcal{T}(.)$ is the nonlinearity applied element-wise to the affine transform of input. We use sigmoid function as the nonlinearity; therefore, $\mathcal{T}(x)=\frac{1}{1+e^{-x}}$. Given the weight matrix $\mathbf{W}_1$ and the bias vector $\mathbf{b}$,  the computational cost for calculation of $\mathbf{x}_{h_1}$ is $\mathcal{O}(MN)$ according to \eqref{eq:1stLayer} that is the same as cost of one iteration of an iterative algorithm for CS recovery problem. In order to keep this computational cost at each layer, the second hidden layer and the output layers are formulated as 
\begin{align}\label{eq:Layers}
\mathbf{x}_{h_2}=\mathcal{T}(\mathbf{W}_2\mathbf{x}_{h_1}+\mathbf{b}_2) \quad\text{and}\quad \mathbf{\hat{x}}=\mathcal{T}(\mathbf{W}_3\mathbf{x}_{h_2}+\mathbf{b}_3).
\end{align}
In \eqref{eq:Layers} $\mathbf{W}_2 \in \mathbf{R}^{M \times N}$ and $\mathbf{b}_2 \in \mathbf{R}^M$ are the weight matrix and bias vector of the second layer, respectively. Similarly, $\mathbf{W}_3 \in \mathbf{R}^{N \times M}$ and $\mathbf{b}_3 \in \mathbf{R}^N$ are the weight matrix and bias vector of the output layer, respectively. We denote the output of the SDA and its set of parameters by $\mathbf{\hat{x}}$ and $\Omega_{\text{L}}=\{\mathbf{W}_1,\mathbf{b}_1,\mathbf{W}_2,\mathbf{b}_2,\mathbf{W}_3,\mathbf{b}_3\}$, respectively. Therefore, we can define the nonlinear mapping $\mathbf{\hat{x}} = \mathcal{M}_{\text{L}}(\mathbf{y},\Omega_{\text{L}})$ and use the mean squared error (MSE) as the loss function for the training set $\mathcal{D}_{\rm train}$
\begin{align}\label{eq:lossFN}
\mathcal{L}(\Omega_{\text{L}})=\frac{1}{l}\sum_{i=1}^l \| \mathcal{M}_{\text{L}}(\mathbf{y}^{(i)},\Omega_{\text{L}})-\mathbf{x}^{(i)}\|_2^2.
\end{align}
We use backpropagation \cite{rumelhart1988learning} algorithm to minimize the loss function defined in \eqref{eq:lossFN}. Figure \ref{fig:SDA_L} shows the SDA structure fed by linear measurements of original signal.

\subsection{SDA + Nonlinear Measurement Paradigm}
\begin{figure}[t!]
\begin{center}
\includegraphics[width= 7cm]{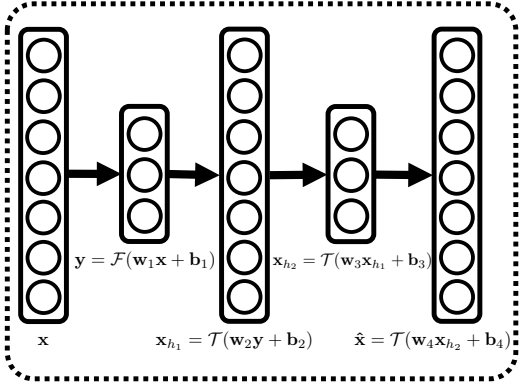}
\caption{SDA for recovering a sparse signal from its non-linear measurements. This is equivalent to having a 4-layer neural network taking non-linear and adaptive measurements from the original signal. The non-linear function used for taking measurements could be different from non-linearity used in other layers. However, it should be analytical tractable in order to fit in backpropagation framework.}
\label{fig:SDA_NL}
\end{center}
\end{figure}

The structure of the SDA for nonlinear measurement paradigm is almost the same as the one in Section \ref{sec:SDAL}. The only difference is that we consider the mapping from original signal to its measurement vector as one layer of the SDA. This extra layer will let SDA to adapt its structure to the training set $\mathcal{D}_{\rm train}$. Therefore, if we have enough data, e.g. lots of natural images (as in ImageNet dataset), we could be hopeful that the measurement matrix is well adapted to the class of signals being under acquisition. We denote this extra layer that is the first layer of the SDA by 
\begin{align}\label{eq:1stNL}
\mathbf{y}=\mathcal{F}(\mathbf{W}_1\mathbf{x}+\mathbf{b}_1),
\end{align}
where $\mathcal{F}(.)$ is the nonlinearity we have used in order to take measurements from the original signal $\mathbf{x}$. $\mathcal{F}(.)$ can be either a sigmoid function used in other layers of the network or other types of nonlinearities or even identity function such that the measurements would be linear and at the same time adapted to the acquired signals just like traditional CS framework. We denote the parameter set of this SDA by $\Omega_{\text{NL}}=\{\mathbf{W}_1,\mathbf{b}_1,\mathbf{W}_2,\mathbf{b}_2,\mathbf{W}_3,\mathbf{b}_3,\mathbf{W}_4,\mathbf{b}_4\}$ and its output by $\mathbf{\hat{x}} = \mathcal{M}_{\text{NL}}(\mathbf{x},\Omega_{\text{NL}})$. The loss function corresponding to this SDA is similar to \eqref{eq:lossFN} with some minor changes
\begin{align}\label{eq:lossNL}
\mathcal{L}(\Omega_{\text{NL}})=\frac{1}{l}\sum_{i=1}^l \|\mathcal{M}_{\text{NL}}(\mathbf{x}^{(i)},\Omega_{\text{NL}})-\mathbf{x}^{(i)}\|_2^2.
\end{align}
Figure \ref{fig:SDA_NL} shows the SDA structure for non-linear measurement paradigm. The next section describes how these SDA structures could be related to the CS recovery problem from the probabilistic point of view.

\section{Probabilistic Relation Between SDA and Compressive Sensing}\label{sec:theory}
In this section we provide a probabilistic interpretation that explains the success of SDA in solving the structured signal recovery problem. As introduced in Section \ref{sec:arc}, the deep network that we are using to solve this problem is basically stacked version of denoising autoencoders. At the first layer, this deep network is fed by a training example that is either compressed measurements of an image in training set (Figure \ref{fig:SDA_L}) or the original image itself (Figure \ref{fig:SDA_NL}). The next layers are then fed by the latent representation (or output code) of the denoising autoencoder found on their corresponding previous layer. We perform an unsupervised pre-training on this deep architecture that is justified in \cite{bengio2007greedy}. The authors in \cite{bengio2007greedy} have explained how unsupervised pre-training helps the corresponding optimization problem in deep networks by initializing the weights in all layers in a region near a good local minimum of loss function.

In the stacked version of denoising autoencoders, the unsupervised pre-training phase is done one layer at a time. Each layer of this deep network is pre-trained as a denoising autoencoder. In other words, it is trained by minimizing the error in reconstructing its input (that is the output code of the previous layer) from the noisy version of it. As we proceed in the pre-training phase, once the first $t$ layers are trained, we can compute the corresponding latent representation (or output code) of the first $t$-layers and use it as an input in order to train the $t+1$-th layer.

An important aspect of this pre-training phase, is the connection between the Restricted Boltzmann Machine (RBM) \cite{hinton2002training} and denoising autoencoder. RBM is a generative model that can learn probability distribution underlying its input data. It has a set of visible units $\mathbf{v}$ and a set of hidden units $\mathbf{h}$. Since it is a energy-based model \cite{lecun2006tutorial}, it associate a scalar energy $E(\mathbf{v},\mathbf{h})$ to each configuration of the visible and hidden units. The joint probability distribution is defined as $\mathbb{P}(\mathbf{v},\mathbf{h})=\frac{1}{Z}e^{-E(\mathbf{v},\mathbf{h})}$, where $Z$ is called the partition function used for normalizing the probability distribution. Training an RBM is equivalent to configuring its energy function such that desirable configurations have low energy. As an example, the energy function corresponding to Gaussian-Bernoulli RBM \cite{krizhevsky2009learning} with real-valued visible units $\mathbf{v}$ and binary hidden units $\mathbf{h}$ is 
\begin{align}\label{equ:energy}
&E(\mathbf{v},\mathbf{h}|\mathbf{W},\mathbf{b},\mathbf{c}) \\
&=\sum_{i=1}^{n_v}\frac{(v_i-b_i)^2}{2\sigma_i^2}-\sum_{i=1}^{n_v} \sum_{j=1}^{n_h} W_{ij}h_j\frac{v_i}{\sigma_i}-\sum_{j=1}^{n_h}c_j h_j, \nonumber
\end{align}
where $W_{ij}$ denote weights connecting visible and hidden units. $\sigma_i$ is the standard deviation associated with the $i$-th Gaussian visible unit. Finally, $b_i$ and $c_j$ denote biases corresponding to visible and hidden units. Training an RBM in this case is the process of adjusting weights $W_{ij}$ and biases $b_i$ and $c_j$ such that the probability distribution it represents fits the training data as well as possible. 

The authors in \cite{kamyshanska2014potential} have shown the derivation of an energy function for autoencoders by interpreting them as dynamical systems \cite{seung1997learning}. In particular, the authors in \cite{kamyshanska2014potential} and \cite{vincent2011connection} have shown that the energy function of an autoencoder with sigmoidal hidden layer and real-valued observations is identical to the free energy of corresponding RBM with Gaussian visible units and binary hidden units. 

Suppose that we want to train a denoising autoencoder with sigmoidal hidden layer to compress signals from a class of probability distribution. This training is equivalent to learning a set of weights and biases that will result in low energy for signals from that probability distribution. In other words, it is equivalent to adjusting set of weights and biases such that the reconstruction error it has for recovering signals (drawn from that probability distribution) from their compressed representation, is as small as possible. 

As an example, suppose that in our training set $\mathcal{D}_{\rm train} = \{(\mathbf{y}^{(1)},\mathbf{x}^{(1)}),(\mathbf{y}^{(2)},\mathbf{x}^{(2)}),\ldots,(\mathbf{y}^{(l)},\mathbf{x}^{(l)})\}$, the original signals $\mathbf{x}^{(i)}$s are drawn from a probability distribution $\mathcal{P}$. As derived in \cite{kamyshanska2014potential} and \cite{vincent2011connection}, the energy function of a denoising autoencoder with sigmoidal hidden units is 
\begin{align}
E(\mathbf{x})=\sum_{j}\log(1+\exp(W_{.j}^T\mathbf{x}+b_j))-\frac{1}{2}\|\mathbf{x}-\mathbf{c}\|_2^2+{\rm const},
\end{align}
where $W_{ij}$ denote weights connecting visible and hidden units, and $b_j$ and $c_i$ are biases for hidden layer and reconstruction layer. Training a denoising autoencoder based on the training set $\mathcal{D}_{\rm train}$ in this case is the process of adjusting the weights $W_{ij}$ and biases $b_j$ and $c_i$ such that the reconstruction error for any signal drawn from probability distribution $\mathcal{P}$ (and not necessarily in $\mathcal{D}_{\rm train}$) is as small as possible.

Similarly, suppose that we want to train a denoising autoencoder with sigmoidal hidden layer to decompress data that is originally (i.e., before compression) coming from the probability distribution $\mathcal{P}$. In this case, the autoencoder learns a set of weights and biases that will result in low energy for compressed signals drawn originally from the probability distribution $\mathcal{P}$. In other words, training is equivalent to adjusting set of weights and biases such that the reconstruction error it has for recovering compressed signals (drawn originally from the probability distribution $\mathcal{P}$) from their decompressed representation (in hidden layer), is as small as possible. This training will end up in retrieving the original signals (from the probability distribution $\mathcal{P}$) as decompressed representations in hidden layer since they are the origin of compressed data.

 This is fairly similar to the optimization problem in \eqref{eq:l0}. In \eqref{eq:l0} we have the measurement vector (compressed data), we know the original signal model ($k$-sparse), and the goal is to retrieve the original signal from the compressed measurements. The considerable difference though is the fact that in \eqref{eq:l0} we need an optimization algorithm to retrieve the signal from its measurements. However, in an autoencoder (or deep networks in general) we need to pass the compressed data into a trained feedforward network without any need to solve an optimization problem. 

Once we are done with pre-training of all the layers, we perform supervised fine-tuning on the weights and biases of the pre-trained SDA. More precisely, we take the encoding part of each denoising autoencoder, stack them together, and use back-propagation algorithm to minimize the MSE on reconstructing the images from their compressed measurements.

\section{Simulation Results}\label{sec:simul}
In this section we study the performance of our proposed framework for structured signal recovery and compare them with the state-of-the-art results. We first describe the implementation of the proposed models. After that, we compare our method with other CS recovery algorithms. We do this comparison based on both the quality of reconstruction (\text{\sl{PSNR}}) and speed of recovery.
\subsection{Implementation}\label{sec:impl}
Autoencoders are very similar to multilayer perceptron (MLP) in structure. In other words, all the units in input layer of an autoencoder are connected to all the units in hidden layer and similarly all the units in hidden layer are connected to all the units in output layer. Therefore, as image size grows, we have to train a larger network as well. This issue poses a huge computational complexity on running the Backpropagation algorithm in addition to increasing the chance of overfitting. 

\begin{figure}[t!]
\begin{center}
\includegraphics[width= 5.5cm]{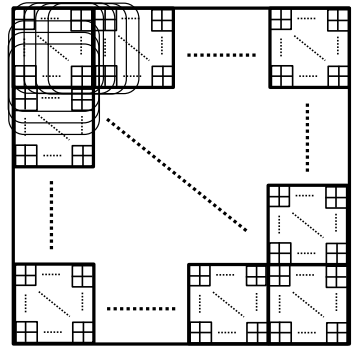}
\caption{Overlapping sub-images of a larger image.}
\label{fig:subimages}
\end{center}
\end{figure}

As a result and instead, we design a neural network for recovering small sub-images (by sub-image we mean a large image patch) instead of a large image. However, this will not block our way to compressively measure and recover large images. We can decompose a large image into several non-overlapping or overlapping sub-images and compressively measure and recover each of them. In the case of non-overlapping sub-images, we will basically have a blocky reconstruction of the original image by putting reconstructed neighbor sub-images beside each other. On the other hand, if we are working with overlapping sub-images, we place each recovered sub-image at its corresponding location in the original image and average on the overlapping sub-images. Figure \ref{fig:subimages} shows a visualization of overlapping sub-images in a larger image. 

In this paper, we have trained our deep neural network based on sub-images of size $32 \times 32$. We used natural images from the ILSVRC 2014 ImageNet dataset \cite{russakovsky2014imagenet} for both training and testing the network. For dataset preparation, we extracted the central $256\times256$ part of each image, turned it into grayscale, and chopped it into $32 \times 32$ sub-images. During the training phase, we did not use overlapping sub-images so as to have a diverse training set. However, during the test phase, we use overlapping sub-images and averaging method as described earlier in this section. We normalized each image pixel value such that they would be between 0 and 1. According to the results in \cite{glorot2010understanding}, we sampled the initial weights of our neural network from a uniform distribution $U\left[ -4\sqrt{\frac{6}{\text{fan}_{\text{in}}+\text{fan}_{\text{out}}}},4\sqrt{\frac{6}{\text{fan}_{\text{in}}+\text{fan}_{\text{out}}}}\right]$, where $\text{fan}_{\text{in}}$ and $\text{fan}_{\text{out}}$ are number of units in input and output layers, respectively. 

As we mentioned earlier, we use denoising autoencoders as building blocks of our network. Therefore, we corrupt the input of each layer with a Gaussian noise having zero mean and standard deviation of 0.2 and let the layer to reconstruct its corrupted input as the pre-training phase. We then use backpropagation algorithm to fine-tune the weights and biases. We implemented our deep neural networks using Theano package \cite{bergstra2010theano} and used GPU on Amazon web service (AWS) platform.

\begin{figure}[t!]
\begin{center}
\includegraphics[width= 9cm]{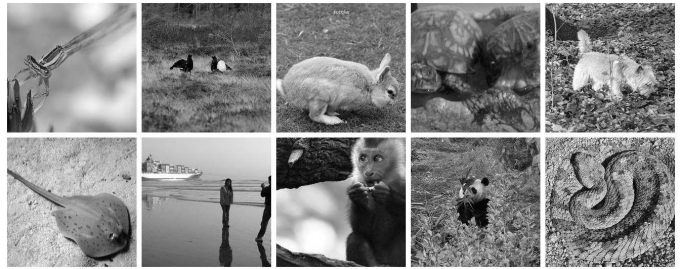}
\caption{Test images for Table \ref{table:PSNR}.}
\label{fig:Images}
\end{center}
\end{figure}

\begin{table}
\begin{center}
\resizebox{\columnwidth}{!}{
\begin{tabular}{|c || c  c  c  c  c c | } 
 \hline
 &  L-SDA &  NL-SDA & D-AMP &  O-NL-SDA  &  Tiled D-AMP & TV\\ [0.5ex] 
 \hline\hline
 \textbf{Damselfly}& \textit{29.01} & 30.32 & \textbf{45.97} & \textit{30.85} & 30.51 & 27.46 \\ 
 \hline
 \textbf{Birds}& 24.93 & \textit{26.19} & 25.58 & \textbf{26.62} & 24.45  & 21.46 \\ 
 \hline
 \textbf{Rabbit}& 25.42 & \textit{26.80} & 26.37& \textbf{27.24} & 25.04& 19.79\\ 
 \hline
 \textbf{Turtle}& 31.07 & 33.79 & \textit{34.17} & \textbf{34.65} & 32.08 & 27.16\\ 
 \hline
 \textbf{Dog}& 19.76 & \textit{21.03} & 19.71& \textbf{21.55} & 18.45& 13.96 \\ 
 \hline
 \textbf{Eagle Ray}& 25.00& \textit{26.18} & 25.37 & \textbf{26.57} & 24.68& 16.30 \\ 
 \hline
 \textbf{Boat}& 29.67& 31.96& \textbf{41.75} & 33.11 & \textit{33.49} & 27.63\\ 
 \hline
\textbf{Monkey}& 28.33& 29.74 & \textbf{34.00} &  \textit{30.32} & 28.66 & 26.47\\ 
\hline
\textbf{Panda} & 19.82& \textit{20.68} & 19.66 & \textbf{21.00} & 18.61 & 18.31\\
\hline
\textbf{Snake} & 16.42 & \textit{17.39} & 16.33& \textbf{17.72} & 15.46 & 10.42 \\ 
\hline
\end{tabular}}
\caption{ Quality of reconstruction ({\em PSNR} in  {\sl dB}) for different images in Figure \ref{fig:Images} and different algorithms. The under-sampling ratio ($\frac{M}{N}$) is assumed to be constant and equal to 0.25 in all the cases.}
\label{table:PSNR}
\end{center}
\end{table}

\subsection{Comparison with Other Methods}
In this section we compare the performance of structured signal recovery using deep neural networks with other recovery algorithms. These other algorithms include
\begin{itemize}
\item  One of the state-of-the-art methods that is the denoising-based approximate message passing (D-AMP) \cite{metzler2014denoising}.
\item The total variation (TV) minimization \cite{candes2006robust} which is famous for its intriguing properties for image recovery.
\item The parameterless approximate message passing (P-AMP) \cite{mousavi2013parameterless} employing sparsity in wavelet domain.
\item the tiled version of D-AMP (Tiled D-AMP).
\end{itemize}
By the Tiled D-AMP we mean D-AMP being applied to the non-overlapping sub-images similar to what we mentioned for the SDA-based methods in Section \ref{sec:impl}. We introduce the Tiled D-AMP for the sake of fairness in here. Since due to huge computational complexity we could not train the SDAs for recovering large images, we wanted to compare the performance of the D-AMP if it is applied in the same way that we apply the SDA-based methods, i.e., recovering images from blocky reconstruction of smaller sub-images. 

Table \ref{table:PSNR} shows the summary of results for recovering 10 different images in Figure \ref{fig:Images}.  As is clear from Table \ref{table:PSNR} there is not an obvious winner among different methods. In some cases, the L-SDA (SDA + Linear Measurements) and NL-SDA (SDA + Non-linear Measurements) and O-NL-SDA (SDA + Overlapping Non-linear Measurements) have better performance in comparison with the D-AMP and min-TV. 

More specifically and according to our simulation results, whenever the acquired image has an irregular structure such that there are not to many similar patches in the image, then the L-SDA, NL-SDA, and O-NL-SDA have better performance comparing to the D-AMP and min-TV. 

For example, in Figure \ref{fig:Images} the dog and panda images have irregular structure and texture. As we can see in Table \ref{table:PSNR} for these images, all the recovery methods based on deep learning have a better performance in comparison with the D-AMP which is one of the state-of-the-art CS recovery algorithms. On the other hand, the damselfly and boat images have very smooth and regular structure and hence lots of similar patches. As a result, we see from Table \ref{table:PSNR} that the D-AMP is outperforming methods based on deep learning since the D-AMP is utilizing these similar patches to enhance the image reconstruction. 

One more interesting point about the Table \ref{table:PSNR} is that in all cases except one, the NL-SDA and O-NL-SDA are outperforming the Tiled D-AMP. This fact shows the potential of deep learning and the fact that if we were able to train huge networks in a reasonable time or coming up with another network structure, we might be able to outperform the D-AMP in almost all the cases. We leave this problem as an avenue for future work. 

\begin{figure}[t!]
\begin{center}
\includegraphics[width= 9cm]{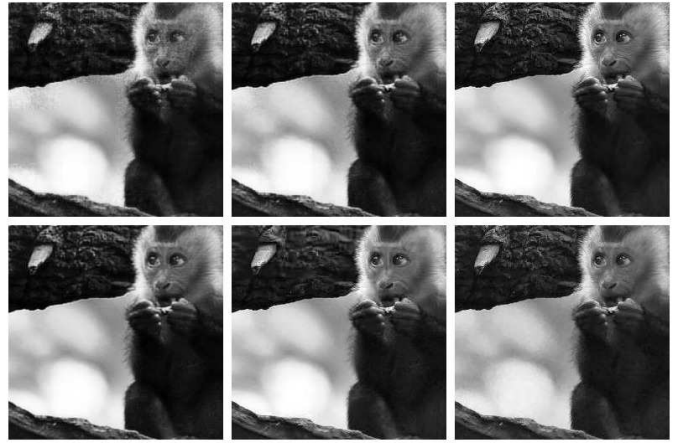}
\caption{Reconstructed monkey image using different algorithms with $\frac{M}{N}=0.4$. Clockwise from upper left: SDA+Linear Measurements ({\sl PSNR=29.96 dB}). SDA+Nonlinear Measurements ({\sl PSNR=31.15 dB}). D-AMP ({\sl PSNR=38.48 dB}). TV ({\sl PSNR=29.67 dB}). Tiled D-AMP ({\sl PSNR=31.56 dB}). Overlapping SDA+Nonlinear Measurements (\sl{PSNR=32 dB}). 
}
\label{fig:reconMonkey}
\end{center}
\end{figure}

\begin{figure}[t!]
\begin{center}
\includegraphics[width= 9cm]{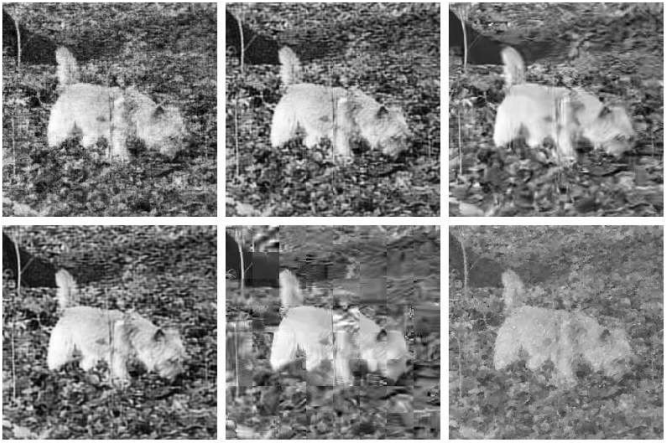}
\caption{Reconstructed dog image using different algorithms with $\frac{M}{N}=0.3$. Clockwise from upper left: SDA+Linear Measurements ({\sl PSNR=20.19 dB}). SDA+Nonlinear Measurements ({\sl PSNR=21.27 dB}). D-AMP ({\sl PSNR=20.25 dB}). TV ({\sl PSNR=14.68 dB}). Tiled D-AMP ({\sl PSNR=18.54 dB}). Overlapping SDA+Nonlinear Measurements ({\sl PSNR=21.88 dB}).}
\label{fig:reconDog}
\end{center}
\end{figure}

\begin{figure}[t!]
\begin{center}
\includegraphics[width= 9cm]{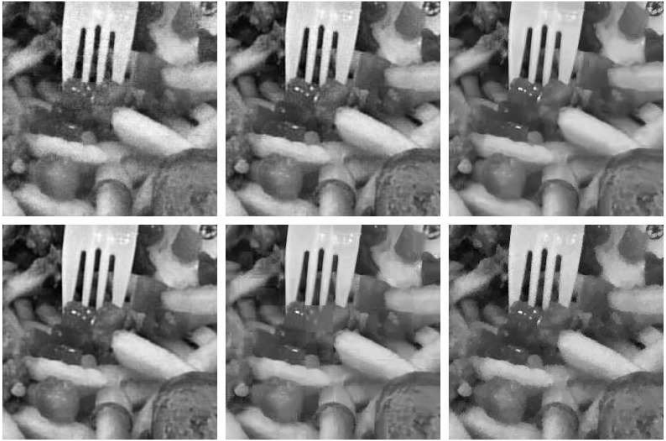}
\caption{Reconstructed Food and Fork image using different algorithms with $\frac{M}{N}=0.25$. Clockwise from upper left: SDA+Linear Measurements ({\sl PSNR=29.45 dB}). SDA+Nonlinear Measurements ({\sl PSNR=31.79 dB}). D-AMP ({\sl PSNR=31.90 dB}). TV ({\sl PSNR=25.16 dB}). Tiled D-AMP ({\sl PSNR=30.57 dB}). Overlapping SDA+Nonlinear Measurements ({\sl PSNR=32.46 dB}).}
\label{fig:reconFork}
\end{center}
\end{figure}

Figures \ref{fig:reconMonkey}, \ref{fig:reconDog}, and \ref{fig:reconFork} show 3 examples of set of reconstructed images. As we can see, in Figure \ref{fig:reconDog} that we do not have a regular and smooth structure, the NL-SDA and O-NL-SDA are outperforming the D-AMP. In Figure \ref{fig:reconFork} that has both regular and irregular textures, we can see that the D-AMP is outperforming the L-SDA and NL-SDA; however, the O-NL-SDA is outperforming the D-AMP. Finally, in Figure \ref{fig:reconMonkey} that mostly has a smooth and regular texture, we see that the D-AMP is outperforming SDA-based methods although the O-NL-SDA is outperforming the Tiled D-AMP.

Although there is not a clear winner among the different methods from the reconstruction quality point of view, our simulation results show that the L-SDA and NL-SDA beat the other methods from the reconstruction time perspective. This is clear both intuitively and mathematically since in all the other methods, we need to solve an optimization problem. We solve this optimization problem by using either convex optimization techniques (e.g. linear programming) or greedy algorithms. However, the L-SDA and NL-SDA do not need to solve any optimization problem. They just use a feed-forward neural network to recover images from their measurements. Table \ref{table:SEC} shows reconstruction time for different recovery algorithms and different under-sampling ratios. As we can see, for under-sampling ratio of 0.4, the NL-SDA and L-SDA are almost 1,000,000 times faster than the D-AMP.

\begin{figure}[t!]
\begin{center}
\includegraphics[width= 9.5cm]{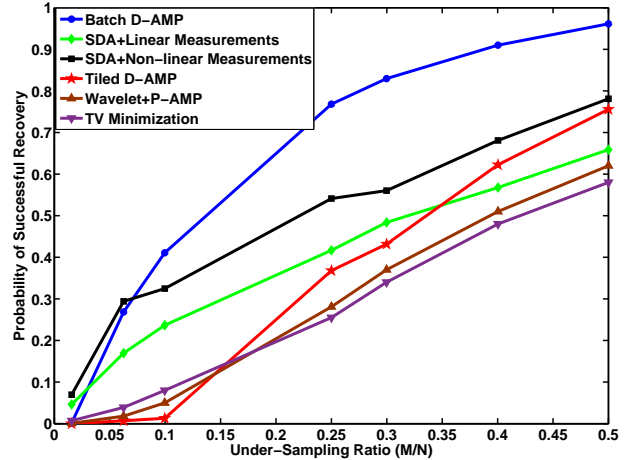}
\caption{Average probability of successful signal recovery for different under-sampling ratios and different algorithms. In order to calculate the probability of successful recovery we have used 1881 Monte Carlo samples. If we denote the original signal by $\mathbf{x}$ and the recovered signal by $\mathbf{\hat{x}}$, then we call a recovery successful if $\frac{\|\mathbf{\hat{x}}-\mathbf{x}\|_2}{\|\mathbf{x}\|_2} \leq 0.01$. }
\label{fig:PT}
\end{center}
\end{figure}

Figure \ref{fig:PT} shows the plot of average probability of successful recovery for different under-sampling ratios and different recovery algorithms. In order to calculate the probability of successful recovery we have used 1881 Monte Carlo samples. For each under-sampling ratio $\delta$ and for the $j$-th Monte Carlo sample, we define the success variable $\varphi_{\delta,j} = \mathbb{I}\left(\frac{\|\mathbf{\hat{x}}^{(j)}-\mathbf{x}^{(j)}\|_2}{\|\mathbf{x}^{(j)}\|_2} \leq 0.01\right)$ where $\mathbf{x}^{(j)}$ is the $j$-th Monte Carlo sample, $\mathbf{\hat{x}}^{(j)}$ denotes the corresponding recovered image, and $\mathbb{I}(.)$ denotes the indicator function. We then define the empirical success probability as $P_{\delta} = \frac{1}{s}\sum_{j=1}^{s}\varphi_{\delta,j}$. As we can see in Figure \ref{fig:PT}, for small under-sampling ratios (less than 0.06), SDA-bases methods are outperforming the D-AMP. In addition, for larger under-sampling ratios the D-AMP is outperforming SDA-based methods. Nevertheless, SDA-based methods (specially the NL-SDA) are outperforming the Tiled D-AMP, P-AMP (employing sparsity in wavelet domain), and TV minimization in a large range of under-sampling ratios. Of course, comparing SDA-based methods with the Tiled D-AMP is a fairer comparison rather than comparing them with D-AMP alone.

\begin{figure}[t!]
\begin{center}
\includegraphics[width= 9.5cm]{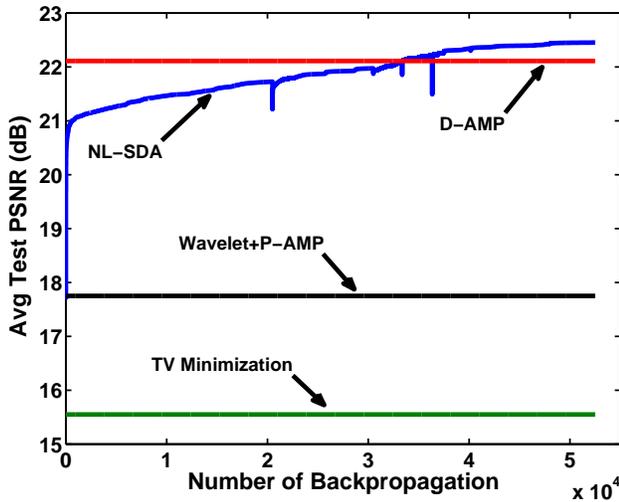}
\caption{Convergence of Backpropagation over different iterations and comparing the test result with other methods. In this plot average PSNR is calculated over 1881 test images each acquired with under-sampling ratio of 0.6, i.e., $\frac{M}{N}=0.06$.}
\label{fig:bp}
\end{center}
\end{figure}

Finally, Figure \ref{fig:bp} denotes the convergence curve of fine tuning step in training our deep neural network. It shows the average PSNR (in dB) on test images over different iterations of backpropagation algorithm. This figure shows that for the under-sampling ratio of 0.06, the NL-SDA method has started to outperform the D-AMP method after $3.5\times10^4$ running of backpropagation algorithm. The jumps in this plot is due to feeding the neural network with new training data.

\section{Conclusion and Future Work}\label{sec:conclu}

In this work, we developed a new framework for sensing and recovering structured signals. This framework is able to learn a structured representation from training data, support both linear and mildly nonlinear measurements, and efficiently computes a signal estimate. In particular, we used a stacked version of denoising autoencoders, as an unsupervised feature learner. We showed how SDA enables us to capture statistical dependencies between the different elements of certain signals and improve signal recovery performance as compared to the CS approach. 

We should note that GRBMs treat different components of the input image vector as conditionally independent given the hidden layer state. This is an important limitation in modeling natural images using GRBM (and denoising autoencoders correspondingly). One important direction for future work is to come up with a model that can easily be extended to large images. In addition, it should capture relationships between pixel intensities rather than assuming them independent conditioned on the hidden layer. This model will let us to outperform the D-AMP algorithm in the cases that SDA-based methods introduced in this paper were not able to.

\begin{table}
\begin{center}
\resizebox{\columnwidth}{!}{
\begin{tabular}{|c || c  c  c  c  c c| } 
 \hline
 $\frac{M}{N}$&    L-SDA &  NL-SDA & D-AMP &  O-NL-SDA &  Tiled D-AMP &  TV \\ [0.5ex] 
  \hline\hline
 0.06& \textbf{0.002} & \textbf{0.002} & 74.79 & 1.01& 43.99& 45.94 \\ 
 \hline
 0.1& \textbf{0.002 } & \textbf{0.002 } & 92.21 & 1.03  & 50.00 & 39.82 \\ 
 \hline
 0.25 & \textbf{0.002 } & \textbf{0.002 } & 108.61 & 1.07 & 50.84  & 43.33\\ 
 \hline
 0.4 & \textbf{0.002 } & \textbf{0.002 } & 1900.68  & 1.22 & 39.55  & 38.83  \\ 
 \hline
 
\end{tabular}}
\caption{Running time (in sec.) for recovering the dog image for different under-sampling ratios and different algorithms. Numbers with bold face show the winner in each row.}
\label{table:SEC}
\end{center}
\end{table}

\bibliographystyle{IEEETran}
\bibliography{Allerton_v1}

\begin{thebibliography}{10}
\providecommand{\url}[1]{#1}
\csname url@samestyle\endcsname
\providecommand{\newblock}{\relax}
\providecommand{\bibinfo}[2]{#2}
\providecommand{\BIBentrySTDinterwordspacing}{\spaceskip=0pt\relax}
\providecommand{\BIBentryALTinterwordstretchfactor}{4}
\providecommand{\BIBentryALTinterwordspacing}{\spaceskip=\fontdimen2\font plus
\BIBentryALTinterwordstretchfactor\fontdimen3\font minus
  \fontdimen4\font\relax}
\providecommand{\BIBforeignlanguage}[2]{{%
\expandafter\ifx\csname l@#1\endcsname\relax
\typeout{** WARNING: IEEEtran.bst: No hyphenation pattern has been}%
\typeout{** loaded for the language `#1'. Using the pattern for}%
\typeout{** the default language instead.}%
\else
\language=\csname l@#1\endcsname
\fi
#2}}
\providecommand{\BIBdecl}{\relax}
\BIBdecl

\bibitem{donoho2006compressed}
D.~L. Donoho, ``Compressed sensing,'' \emph{Information Theory, IEEE
  Transactions on}, vol.~52, no.~4, pp. 1289--1306, 2006.

\bibitem{baraniuk2007compressive}
R.~G. Baraniuk, ``Compressive sensing,'' \emph{IEEE Signal Processing
  Magazine}, vol.~24, no.~4, 2007.

\bibitem{candes2006compressive}
E.~J. Cand{\`e}s, ``Compressive sampling,'' in \emph{Proceedings of the
  International Congress of Mathematicians}, vol.~3.\hskip 1em plus 0.5em minus
  0.4em\relax Madrid, Spain, 2006, pp. 1433--1452.

\bibitem{hinton2006reducing}
G.~E. Hinton and R.~R. Salakhutdinov, ``Reducing the dimensionality of data
  with neural networks,'' \emph{Science}, vol. 313, no. 5786, pp. 504--507,
  2006.

\bibitem{baron2005distributed}
D.~Baron, M.~B. Wakin, M.~F. Duarte, S.~Sarvotham, and R.~G. Baraniuk,
  ``Distributed compressed sensing.''

\bibitem{candes2006near}
E.~J. Cand{\`e}s and T.~Tao, ``Near-optimal signal recovery from random
  projections: Universal encoding strategies?'' \emph{Information Theory, IEEE
  Transactions on}, vol.~52, no.~12, pp. 5406--5425, 2006.

\bibitem{donoho2009message}
D.~L. Donoho, A.~Maleki, and A.~Montanari, ``Message-passing algorithms for
  compressed sensing,'' \emph{Proceedings of the National Academy of Sciences},
  vol. 106, no.~45, pp. 18\,914--18\,919, 2009.

\bibitem{needell2009cosamp}
D.~Needell and J.~A. Tropp, ``Cosamp: Iterative signal recovery from incomplete
  and inaccurate samples,'' \emph{Applied and Computational Harmonic Analysis},
  vol.~26, no.~3, pp. 301--321, 2009.

\bibitem{blumensath2009iterative}
T.~Blumensath and M.~E. Davies, ``Iterative hard thresholding for compressed
  sensing,'' \emph{Applied and Computational Harmonic Analysis}, vol.~27,
  no.~3, pp. 265--274, 2009.

\bibitem{beck2009fast}
A.~Beck and M.~Teboulle, ``A fast iterative shrinkage-thresholding algorithm
  for linear inverse problems,'' \emph{SIAM Journal on Imaging Sciences},
  vol.~2, no.~1, pp. 183--202, 2009.

\bibitem{candes2008restricted}
E.~J. Cand{\`e}s, ``The restricted isometry property and its implications for
  compressed sensing,'' \emph{Comptes Rendus Mathematique}, vol. 346, no.~9,
  pp. 589--592, 2008.

\bibitem{malloy2014near}
M.~L. Malloy and R.~D. Nowak, ``Near-optimal adaptive compressed sensing,''
  \emph{Information Theory, IEEE Transactions on}, vol.~60, no.~7, pp.
  4001--4012, 2014.

\bibitem{haupt2009adaptive}
J.~Haupt, R.~Nowak, and R.~Castro, ``Adaptive sensing for sparse signal
  recovery,'' in \emph{Digital Signal Processing Workshop and 5th IEEE Signal
  Processing Education Workshop, 2009. DSP/SPE 2009. IEEE 13th}, 2009, pp.
  702--707.

\bibitem{haupt2012sequentially}
J.~Haupt, R.~Baraniuk, R.~Castro, and R.~Nowak, ``Sequentially designed
  compressed sensing,'' in \emph{Proc. IEEE Work. Stat. Signal Processing},
  2012, pp. 401--404.

\bibitem{haupt2009compressive}
J.~D. Haupt, R.~G. Baraniuk, R.~M. Castro, and R.~D. Nowak, ``Compressive
  distilled sensing: Sparse recovery using adaptivity in compressive
  measurements,'' in \emph{Proc. Asilomar Conf. Signals, Systems, and
  Computers}, 2009, pp. 1551--1555.

\bibitem{mallat1999wavelet}
S.~Mallat, \emph{A wavelet tour of signal processing}.\hskip 1em plus 0.5em
  minus 0.4em\relax Academic Press, 1999.

\bibitem{crouse1998wavelet}
M.~S. Crouse, R.~D. Nowak, and R.~G. Baraniuk, ``Wavelet-based statistical
  signal processing using hidden markov models,'' \emph{Signal Processing, IEEE
  Transactions on}, vol.~46, no.~4, pp. 886--902, 1998.

\bibitem{de1997non}
J.~S. De~Bonet and P.~A. Viola, ``A non-parametric multi-scale statistical
  model for natural images,'' in \emph{Proc. Adv. in Neural Processing Systems
  (NIPS)}, 1997.

\bibitem{aharon2006img}
M.~Aharon, M.~Elad, and A.~Bruckstein, ``K-svd: An algorithm for designing
  overcomplete dictionaries for sparse representation,'' \emph{Signal
  Processing, IEEE Transactions on}, vol.~54, no.~11, pp. 4311--4322, 2006.

\bibitem{burger2012image}
H.~C. Burger, C.~J. Schuler, and S.~Harmeling, ``Image denoising: Can plain
  neural networks compete with {BM3D}?'' in \emph{Proc. IEEE Int. Conf. Comp.
  Vision, and Pattern Recognition (CVPR)}, 2012, pp. 2392--2399.

\bibitem{eigen2013restoring}
D.~Eigen, D.~Krishnan, and R.~Fergus, ``Restoring an image taken through a
  window covered with dirt or rain,'' in \emph{Proc. IEEE Int. Conf. Comp.
  Vision (ICCV)}, 2013, pp. 633--640.

\bibitem{dong2014learning}
C.~Dong, C.~C. Loy, K.~He, and X.~Tang, ``Learning a deep convolutional network
  for image super-resolution,'' in \emph{Proc. European Conf. Comp. Vision
  (ECCV)}.\hskip 1em plus 0.5em minus 0.4em\relax Springer, 2014, pp. 184--199.

\bibitem{rumelhart1988learning}
D.~E. Rumelhart, G.~E. Hinton, and R.~J. Williams, ``Learning representations
  by back-propagating errors,'' \emph{Cognitive Modeling}, vol.~5, p.~3, 1988.

\bibitem{bengio2007greedy}
Y.~Bengio, P.~Lamblin, D.~Popovici, and H.~Larochelle, ``Greedy layer-wise
  training of deep networks,'' \emph{Proc. Adv. in Neural Processing Systems
  (NIPS)}, vol.~19, pp. 153--160, 2007.

\bibitem{hinton2002training}
G.~E. Hinton, ``Training products of experts by minimizing contrastive
  divergence,'' \emph{Neural Computation}, vol.~14, no.~8, pp. 1771--1800,
  2002.

\bibitem{lecun2006tutorial}
Y.~LeCun, S.~Chopra, R.~Hadsell, M.~Ranzato, and F.~Huang, ``A tutorial on
  energy-based learning,'' \emph{Predicting Structured Data}, vol.~1, p.~0,
  2006.

\bibitem{krizhevsky2009learning}
A.~Krizhevsky and G.~Hinton, ``Learning multiple layers of features from tiny
  images,'' 2009.

\bibitem{kamyshanska2014potential}
H.~Kamyshanska and R.~Memisevic, ``The potential energy of an autoencoder,''
  \emph{Pattern Analysis and Machine Intelligence, IEEE Transactions on},
  vol.~37, no.~6, pp. 1261--1273.

\bibitem{seung1997learning}
H.~S. Seung, ``Learning continuous attractors in recurrent networks.'' in
  \emph{Proc. Adv. in Neural Processing Systems (NIPS)}, vol.~97, 1997, pp.
  654--660.

\bibitem{vincent2011connection}
P.~Vincent, ``A connection between score matching and denoising autoencoders,''
  \emph{Neural Computation}, vol.~23, no.~7, pp. 1661--1674, 2011.

\bibitem{russakovsky2014imagenet}
O.~Russakovsky, J.~Deng, H.~Su, J.~Krause, S.~Satheesh, S.~Ma, Z.~Huang,
  A.~Karpathy, A.~Khosla, and M.~Bernstein, ``Imagenet large scale visual
  recognition challenge,'' \emph{International Journal of Computer Vision}, pp.
  1--42, 2014.

\bibitem{glorot2010understanding}
X.~Glorot and Y.~Bengio, ``Understanding the difficulty of training deep
  feedforward neural networks,'' in \emph{International Conference on
  Artificial Intelligence and Statistics}, 2010, pp. 249--256.

\bibitem{bergstra2010theano}
J.~Bergstra, O.~Breuleux, F.~Bastien, P.~Lamblin, R.~Pascanu, G.~Desjardins,
  J.~Turian, D.~Warde-Farley, and Y.~Bengio, ``Theano: a cpu and gpu math
  expression compiler,'' in \emph{Proceedings of the Python for Scientific
  Computing Conference (SciPy)}, vol.~4.\hskip 1em plus 0.5em minus 0.4em\relax
  Austin, TX, 2010, p.~3.

\bibitem{metzler2014denoising}
C.~A. Metzler, A.~Maleki, and R.~G. Baraniuk, ``From denoising to compressed
  sensing,'' \emph{arXiv preprint arXiv:1406.4175}, 2014.

\bibitem{candes2006robust}
E.~Cand\`es, J.~Romberg, and T.~Tao, ``Robust uncertainty principles: Exact
  signal reconstruction from highly incomplete frequency information,''
  \emph{IEEE Trans. Inform. Theory}, vol.~52, no.~2, pp. 489--509, February
  2006.

\bibitem{mousavi2013parameterless}
A.~Mousavi, A.~Maleki, and R.~G. Baraniuk, ``Parameterless optimal approximate
  message passing,'' \emph{arXiv preprint arXiv:1311.0035}, 2013.

\end{thebibliography}

\end{document}